\pdfoutput=1

\documentclass[11pt]{article}

\usepackage[final]{acl}

\usepackage{times}
\usepackage{latexsym}

\usepackage[T1]{fontenc}

\usepackage[utf8]{inputenc}

\usepackage{microtype}

\usepackage{inconsolata}

\usepackage{subcaption}
\usepackage{graphicx}
\usepackage{float}

\usepackage{multicol}
\usepackage{multirow}
\usepackage{booktabs}
\usepackage{makecell}
\usepackage{subcaption}

\usepackage{amsmath}
\usepackage{cleveref}

\usepackage{todonotes}

\usepackage[table]{xcolor}

\definecolor{fulloverlap}{HTML}{39761d}
\definecolor{highsim}{HTML}{bf9100}
\definecolor{lowsim}{HTML}{b45e07}
\definecolor{nooverlap}{HTML}{990100}

\makeatletter
\newcommand*\iftodonotes{\if@todonotes@disabled\expandafter\@secondoftwo\else\expandafter\@firstoftwo\fi}  %
\makeatother

\definecolor{purp}{HTML}{791f87}
\definecolor{highlight}{RGB}{255, 255, 0}
\definecolor{bottlegreen}{rgb}{0.0,0.42,0.31}
\definecolor{bblue}{rgb}{0.0, 0.6, 0.8}

\title{False Friends Are \textit{Not} Foes: \\ Investigating Vocabulary Overlap in Multilingual Language Models
}

\author{Julie Kallini,
  Dan Jurafsky, Christopher Potts, Martijn Bartelds \\
  Stanford University \\[1ex]
  \texttt{kallini@stanford.edu}
}

\begin{document}
\maketitle
\begin{abstract}

Subword tokenizers trained on multilingual corpora naturally produce overlapping tokens across languages.
Does token overlap facilitate cross-lingual transfer or instead introduce interference between languages? Prior work offers mixed evidence, partly due to varied setups and confounders, such as token frequency or subword segmentation granularity. To address this question, we devise a controlled experiment where we train bilingual autoregressive models on multiple language pairs under systematically varied vocabulary overlap settings. Crucially, we explore a new dimension to understanding how overlap affects transfer: the semantic similarity of tokens shared across languages. We first analyze our models' hidden representations and find that overlap \emph{of any kind} creates embedding spaces that capture cross-lingual semantic relationships, while this effect is much weaker in models with disjoint vocabularies.
On XNLI and XQuAD, we find that models with overlap outperform models with disjoint vocabularies, and that transfer performance generally improves as overlap increases.
Overall, our findings highlight the advantages of token overlap in multilingual models and show that substantial shared vocabulary remains a beneficial design choice for multilingual tokenizers.
\newline
\newline
\hspace{.5em}\includegraphics[width=1.25em,height=1.25em]{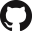}\hspace{.75em}\parbox{\dimexpr\linewidth-2\fboxsep-2\fboxrule}{\url{https://github.com/jkallini/false-friends}}

\end{abstract}

\everypar{\looseness=-1}

\section{Introduction}

Multilingual tokenizers are commonly trained on the concatenation of corpora from multiple languages~\cite{conneau-etal-2020-unsupervised, xue-etal-2021-mt5}, resulting in subword vocabularies with naturally overlapping tokens across languages. While some of these shared tokens may correspond to semantically aligned units across languages (e.g., cognates, named entities), others may arise from coincidental overlaps or have different meanings (e.g., false friends).
Although prior work has demonstrated that token overlap can enhance zero-shot cross-lingual transfer~\citep{pires-etal-2019-multilingual, conneau-etal-2020-emerging}, others report adverse effects depending on the end task~(e.g.,~\citealp{limisiewicz-etal-2023-tokenization}). Some tokenization approaches have aimed to reduce overlap altogether~\cite{chung-etal-2020-improving, liang-etal-2023-xlm}.
These contradictory studies lead us to ask:
\textit{when and how does the presence of overlapping tokens improve cross-lingual transfer?}

\begin{figure}[]
  \centering
  \includegraphics[width=\columnwidth]{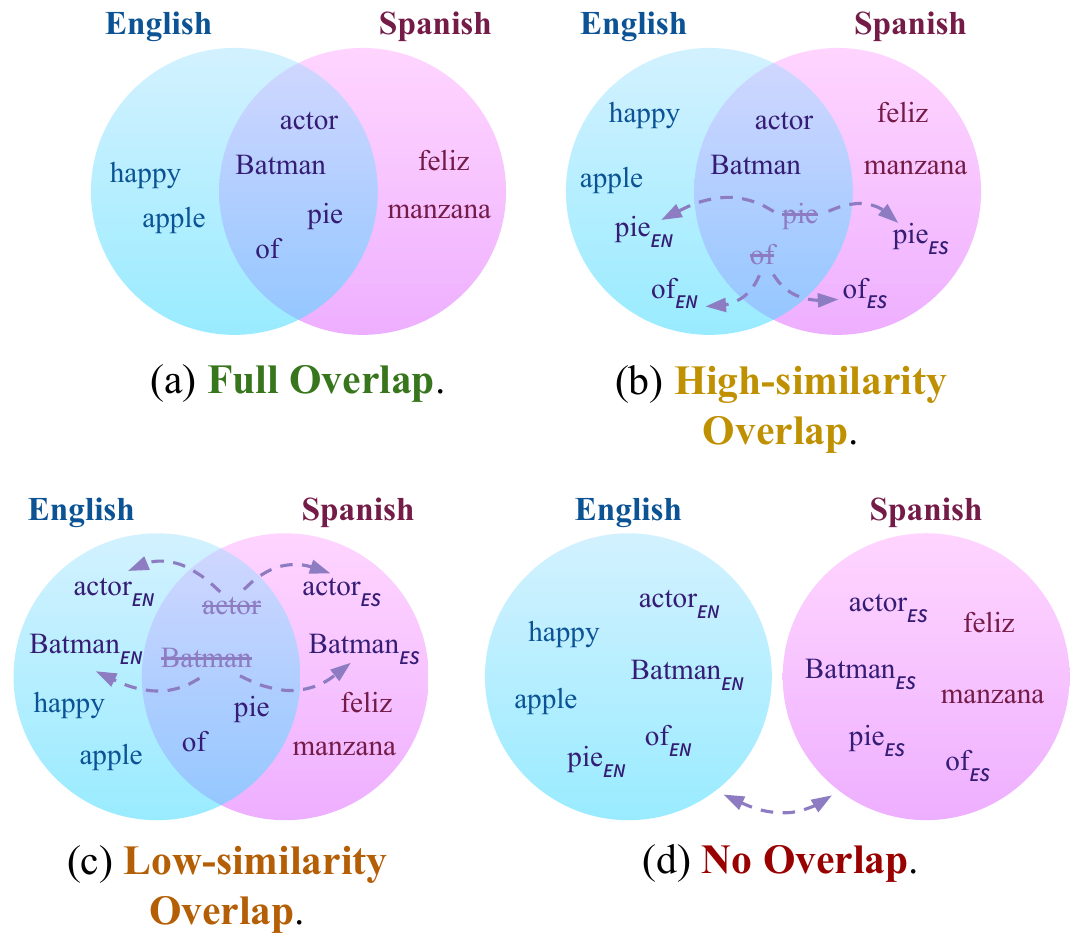}
  \caption{
A visualization of the four overlap settings used in our experiments.
(a) \textcolor{fulloverlap}{\textbf{Full Overlap}}: the two languages share the original tokenizer's native overlapping subwords. These include true cognates and named entities (e.g., \emph{actor}, \emph{Batman}) as well as false cognates or coincidental overlaps (e.g., \emph{pie}, \emph{of}).
(b) \textcolor{highsim}{\textbf{High-similarity Overlap}}: only tokens with the highest cross‑lingual semantic similarity are shared.
(c) \textcolor{lowsim}{\textbf{Low-similarity Overlap}}: only tokens with the lowest cross‑lingual semantic similarity are shared.
(d) \textcolor{nooverlap}{\textbf{No Overlap}}: the two languages' vocabularies are completely disjoint.
  }
  \label{fig:four-settings}
\end{figure}

We answer this question by training bilingual autoregressive models on data from six language pairs, each under four controlled vocabulary overlap settings (\Cref{fig:four-settings}).
In contrast to prior work, we distinguish different types of overlap based on semantic similarity of the tokens in the two languages, as semantic alignment has been shown to impact cross-lingual transfer \cite{Cao2020Multilingual, deshpande-etal-2022-bert, hua-etal-2024-mothello}, while holding subword segmentation granularity and token frequency distributions fixed.
Within pre-trained models, we find that token overlap enables the embedding spaces of the two languages to capture cross-lingual semantic relationships---an effect that is substantially weaker in models with disjoint vocabularies.
When testing zero-shot transfer between languages on the XNLI and XQuAD downstream tasks, models with any amount of overlap consistently outperform models with no overlap, and transfer performance generally improves as overlap increases. We find that tokens with shared meanings across languages contribute most to transfer performance, though any overlap is beneficial.
Our findings offer practical guidance on the design of future multilingual tokenizers.
\looseness=-1

\section{Background and Related Work}

\paragraph{Vocabulary Overlap.} Research on cross-lingual transfer has revealed both advantages and challenges of subword overlap in multilingual models. On the positive side, prior work showed that token overlap provides moderate gains for zero-shot transfer in multilingual BERT (mBERT) on language understanding tasks \cite{pires-etal-2019-multilingual, wu-dredze-2019-beto, dufter-schutze-2020-identifying}.
\citet{conneau-etal-2020-emerging} more closely examined token overlap using three vocabulary-sharing schemes in bilingual encoders and observed that overlap provided marginal improvements on XNLI, NER, and parsing. \citet{K2020Cross-Lingual} similarly reported minimal performance differences due to wordpiece overlap in bilingual BERT models.

More recently, \citet{limisiewicz-etal-2023-tokenization} found that while overlap can benefit sentence‑level tasks and NER, it may degrade performance on syntactic tasks. Similarly, \citet{zhang-etal-2023} show that multilingual corpora contain unexpectedly high levels of overlap, largely due to code-switching and shared vocabularies, which may help explain cross-lingual transfer in dense retrieval models.
\citet{zhang-etal-2025-tomato} extend overlap by merging subwords with different forms but similar meanings into “semantic tokens,” preserving downstream performance with smaller vocabularies. \citet{hammerl-etal-2025-beyond} show that similarity- or alignment-weighted overlap correlates with cross-lingual transfer across different scripts.
Other related work shows that multilingual tokenizers often over-segment low-resource languages, artificially inflating subword overlap \cite{rust-etal-2021-good, NEURIPS2023_74bb24dc, ahia-etal-2023-languages}. This over-segmentation reduces efficiency and degrades representation quality.

Taken together, these studies paint an unclear picture: while vocabulary overlap can create cross-lingual anchors that facilitate transfer, it may introduce interference across languages that hinders modeling. Moreover, the conditions under which overlap is beneficial remain insufficiently explored.
Unlike prior work, we focus on how the semantic similarity of shared tokens affects performance, while carefully controlling for confounders like subword segmentation granularity and token frequency distributions.

\paragraph{Tokenizer Design.} 
Vocabulary overlap has likewise been a central consideration in tokenizer design.
\citet{chung-etal-2020-improving} and \citet{liang-etal-2023-xlm} use clustering methods to de-emphasize token overlap between lexically distinct languages, citing \citet{K2020Cross-Lingual} for the thesis that overlap is not the principal factor in multilingual model effectiveness. In contrast, \citet{patil-etal-2022-overlap} highlight the importance of overlap for transfer and propose a method to promote token overlap between high‑ and low‑resource languages. At the extreme, byte‑ and character‑level models (e.g., CANINE, \citealp{clark-etal-2022-canine}; ByT5, \citealp{xue-etal-2022-byt5}; MrT5, \citealp{kallini2025mrt}; BLT, \citealp{pagnoni-etal-2025-byte}; H-Net, \citealp{hnets}) eliminate subword tokenization altogether. This maximizes vocabulary overlap but comes at a cost to efficiency, presenting unique engineering challenges. \looseness=-1

\section{Approach: Controlled Overlap Settings}
\label{sec:approach}

To systematically vary the vocabulary overlap between two languages according to our four experimental settings (see \Cref{fig:four-settings}), we denote a base tokenizer $\mathcal{T}$ with vocabulary $V$ of size $N$.
We assume that $V = \{0,1,\dots,N-1\}$,
i.e.\ that each token in $V$ is represented by an integer index from 0 to $N-1$. 
Two languages $L_1$ and $L_2$ have corpora $C_1$ and $C_2$, respectively, which we tokenize using $\mathcal{T}$. Let
$V_1 = \{\text{unique tokens in }C_1\} \subseteq V$ and
$V_2 = \{\text{unique tokens in }C_2\} \subseteq V$.
In other words, $V_1$ and $V_2$ are the individual vocabularies of $L_1$ and $L_2$, respectively.
Thus, when tokenizing $C_1$ and $C_2$, the \emph{native overlap} of $\mathcal{T}$ is the set
$O = V_1 \cap V_2$, and the \emph{effective vocabulary size} of $\mathcal{T}$ is $N_\mathrm{eff} = |V_1| + |V_2| - |O|$.

Given a token sequence $X = [x_1, x_2, \dots, x_n]$, where $x_i\in V$,
from language $\ell\in\{L_1,L_2\}$, we define a modified tokenizer $\mathcal{T}'$ that produces
$
X' = [x_1', x_2', \dots, x_n']$,
where each
\begin{equation*}
\begin{split}
x_i' &= 
\begin{cases}
x_i + N,           & \ell = L_2 \text{ and } x_i\notin O',\\
x_i,                 & \text{otherwise.}
\end{cases}
\end{split}
\end{equation*}
Here, $O'\subseteq O$ denotes the subset of tokens we choose to share under a given setting: for $L_1$, all tokens remain unchanged, and for $L_2$, tokens in $O'$ remain unchanged while all others are offset by $N$. This guarantees that $L_1$ and $L_2$ only share $O'$, and $\mathcal{T}'$ has a new effective vocabulary size $N_\mathrm{eff}' = |V_1| + |V_2| - |O'|$. The four choices of $O'$ define our four settings, listed below.

\paragraph{\textcolor{fulloverlap}{Full Overlap}.}  $O' = O$.  Since this only renames certain tokens $x_i \notin O$ from $L_2$, $\mathcal{T}'$ is behaviorally equivalent to $\mathcal{T}$, and $N_\mathrm{eff}' =|V_1| + |V_2| - |O|$.

\paragraph{\textcolor{highsim}{High-similarity Overlap}.} $O' = O_{\mathrm{hi}}$, where $O_{\mathrm{hi}} \subseteq O$ is the set of tokens whose meanings align closely between $L_1$ and $L_2$.  Only these tokens remain shared, so $N_\mathrm{eff}' = |V_1| + |V_2| - |O_{\mathrm{hi}}|$.

\paragraph{\textcolor{lowsim}{Low-similarity Overlap}.} $O' = O_{\mathrm{lo}}$, where $O_{\mathrm{lo}} \subseteq O$ is the set of tokens whose meanings differ across $L_1$ and $L_2$. Only these tokens remain shared, so $N_\mathrm{eff}' = |V_1| + |V_2| - |O_{\mathrm{lo}}|$.

\paragraph{\textcolor{nooverlap}{No Overlap}.} $O' = \emptyset$. Since no tokens are shared, $N_\mathrm{eff}' = |V_1| + |V_2|$.

\vspace{6pt}
\noindent
The details for the semantic partitioning of $O$ into $O_{\mathrm{hi}}$ and $O_{\mathrm{lo}}$ are presented in the next section.

\section{Implementation Details}

\paragraph{Datasets.}

We use CCMatrix \cite{schwenk-etal-2021-ccmatrix}, a large collection of high-quality web-mined parallel texts, for bilingual model pre-training. This allows us to control for the content and the approximate quantity of data in each language. We train on six language pairs: English--Spanish, English--German, English--Turkish, English--Chinese, English--Arabic, and English--Swahili.
English is included in every pair to reflect realistic training scenarios, as English is typically the dominant language in multilingual pre-training datasets.
The second language is selected to cover various language families, scripts, and typological distances from English.
The pre-training corpus for each pair is constructed by shuffling and interleaving sentences from both languages.

\paragraph{Tokenizer and Overlap Partitioning.}
Our base tokenizer $\mathcal{T}$ is the multilingual XLM-R tokenizer~\cite{conneau-etal-2020-unsupervised}, which uses SentencePiece~\cite{kudo-richardson-2018-sentencepiece} with a unigram LM~\cite{kudo-2018-subword}. We found that it offers more effective compression across languages than other tokenizers (see~\Cref{app:tokenizer-compression}).
To divide the native overlap $O$ into high- and low-similarity subsets ($O_{\mathrm{hi}}$ and $O_{\mathrm{lo}}$), we rank tokens in $O$ by their semantic similarity across languages.
For each token $t \in O$, we extract 100 occurrences from $C_1$ (the CCMatrix corpus for $L_1$), pass the sentences through XLM-R, and mean-pool the layer-$l$ contextual embeddings of $t$ to obtain a static embedding $e_1$ (following \citealp{bommasani-etal-2020-interpreting}). The layer $l$ is pre-determined by a sweep we conducted on sets of cognates and non-cognates, as detailed in~\Cref{app:layer-selection}. We repeat this for $C_2$ (the corpus for $L_2$) to compute $e_2$. The cosine similarity between $e_1$ and $e_2$ serves as the token's cross-lingual similarity score.
We rank tokens in $O$ by these scores, assigning the top half to $O_{\mathrm{hi}}$ and the bottom half to $O_{\mathrm{lo}}$. For detailed corpus statistics and overlap metrics for each setting, refer to~\Cref{app:overlap-metrics}.

\paragraph{Models.}
For each language pair and vocabulary setting, we pre-train a separate model, resulting in 24 bilingual models in total. All models are autoregressive Transformers~\cite{vaswani2017attention} with 85M non-embedding parameters, equivalent in size to GPT-2 Small \cite{radford2019languagemodels}.
We train these models due to their architectural similarity with modern LLMs.
See~\Cref{app:pre-training} for additional architecture and optimization details. 

\section{Embedding Similarity Analysis}
\label{sec:analysis}

\begin{figure*}[ht]
  \centering
  \includegraphics[width=1.0\linewidth]{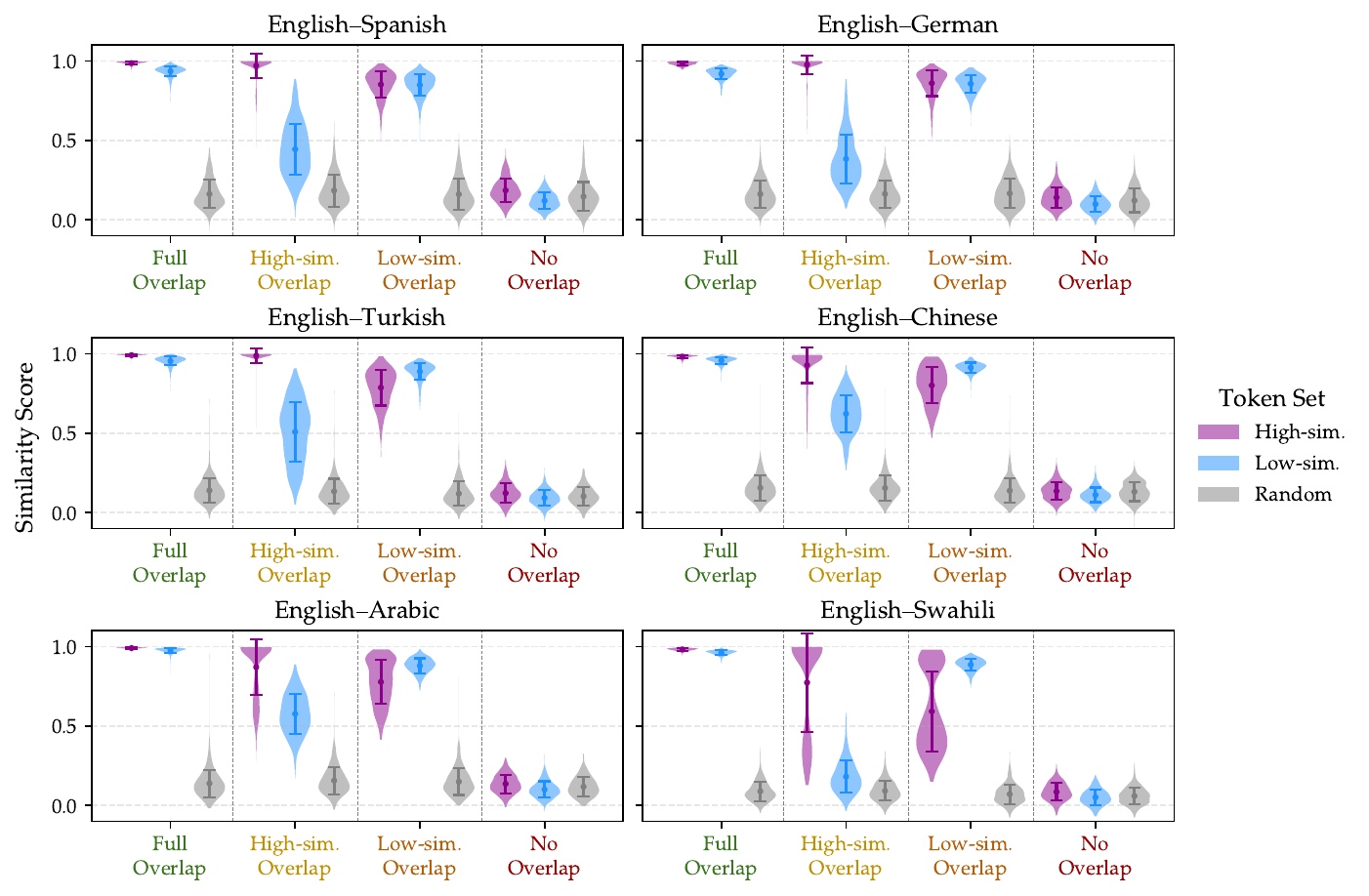}
  \caption{Embedding similarity analysis on pre-trained models for each language pair and vocabulary setting. Cosine similarity is used to measure similarity of tokens in $L_1$ and $L_2$ for a given language pair. The high-sim.\ token set (purple) should have similar meanings; the low-sim.\ token set (blue) should have dissimilar meanings; the random token set (gray) should not share form or meaning, and are shown as a control for anisotropy.}
  \label{fig:xlmr_all_similarity_scores}
\end{figure*}

As a first step in analyzing our pre-trained models, we test how sharing semantically similar or dissimilar tokens influences the model’s learned representations.
We take the 500 most and least similar overlapping tokens for each language pair, ranked using XLM-R as described in the previous section. From a middle layer ($l=6$) of our own models, we extract contextual embeddings for each token to construct a single static embedding of the token for each language using the same method as before. We then ask whether models learn more similar representations for high-similarity tokens and more distinct ones for low-similarity tokens. Crucially, whether these tokens are shared depends on the overlap setting: in the \emph{High-similarity Overlap} condition, the top 500 tokens are shared; in the \emph{Low-similarity Overlap} condition, the bottom 500 are shared.
To control for the high baseline cosine similarities observed in Transformer embeddings due to anisotropy \cite{ethayarajh-2019-contextual}, we additionally measure similarity scores for 500 randomly selected non-overlapping token pairs. 

\paragraph{Results.}

\Cref{fig:xlmr_all_similarity_scores} summarizes the results across all language pairs. With the exception of the \textit{Low-similarity Overlap} setting for English--Spanish and English--German, the difference between the high- and low-similarity token sets was statistically significant for every language pair and overlap condition (unpaired $t$-tests, all Bonferroni-corrected $p < .05$). The effect size (Cohen's $d$) varied with the overlap condition (see~\Cref{tab:cohens-d} for all effect sizes). In the \textit{Full Overlap} and \textit{High-similarity Overlap} settings, high-similarity tokens consistently scored higher than low-similarity tokens, yielding very large effects ($d \in [1.3, 5.1]$). Even in the \textit{No Overlap} setting, the high-similarity tokens scored higher than low-similarity tokens, though effect sizes were smaller ($d \in [0.5, 1.0]$), suggesting that some degree of semantic alignment persists even without shared lexical anchors.  

In contrast, the \textit{Low-similarity Overlap} setting revealed a split in results based on language family. For the closely related language pairs English--Spanish and English--German, no significant differences were observed (Bonferroni-corrected $p = 1$, $d \approx 0$). However, for more typologically distant language pairs (English--Turkish, English--Chinese, English--Arabic, English--Swahili), the effect reversed: low-similarity tokens scored higher than high-similarity tokens, with large negative effect sizes ($d \in [-1.6, -1.0]$). These reversals indicate that in the \textit{Low-similarity Overlap} setting, tokens that do not share meaning become aligned in the embedding space simply because they are shared in the vocabulary, producing misleading or inverted similarity effects. This demonstrates that the \textit{type} of overlap---whether it links semantically similar or dissimilar tokens---critically shapes how cross-lingual models align token representations. Effects are especially pronounced for more distant language pairs, where there is less contextual signal available to counteract the bias introduced by shared but semantically unrelated tokens.

\section{Downstream Task Performance}
\label{sec:downstream-tasks}

We further fine-tune and evaluate our models on two downstream tasks, namely, natural language inference (NLI) and question answering (QA), in a standard zero-shot transfer setup.
For NLI, we train on English MultiNLI \cite{williams-etal-2018-broad} and evaluate on XNLI \cite{conneau-etal-2018-xnli}. For QA, we train on English SQuAD \cite{rajpurkar-etal-2016-squad} and evaluate on XQuAD \cite{artetxe-etal-2020-cross}. Fine-tuning hyperparameters and optimization details are provided in~\Cref{app:fine-tuning}.
\looseness=-1

\begin{table}
\centering
\resizebox{\columnwidth}{!}{%
\begin{tabular}{@{} l@{ \hspace{2pt} }lc@{ \hspace{0pt} }ccc @{}}
\toprule
\multirow{2}{*}{\makecell[l]{\textbf{Language}\\\textbf{Pair}}}
  & \multirow{2}{*}{\makecell[l]{\textbf{Overlap}\\\textbf{Setting}}}
  & \multicolumn{2}{c}{\textbf{XNLI Accuracy (\%)}}
  & \multicolumn{2}{c}{\textbf{XQuAD F1 / EM (\%)}} \\
\cmidrule(lr){3-4} \cmidrule(lr){5-6} 
  &
  & Test ($L_1$)
  & Test ($L_2$)
  & Test ($L_1$)
  & Test ($L_2$) \\

\midrule
\multirow{4}{*}{\makecell[l]{English--\\Spanish}}
  & \textcolor{fulloverlap}{Full}      & \textbf{78.78} & \textbf{74.59} & 63.83 / 51.85 & \textbf{52.84 / 36.47} \\
  & \textcolor{highsim}{High-sim.}     & \textbf{78.52} & \textbf{73.99} & 63.42 / 53.03 & \textbf{48.60 / 31.85} \\
  & \textcolor{lowsim}{Low-sim.}       & \textbf{79.18} & \textbf{74.55} & 63.52 / 51.93 & \textbf{51.57 / 36.13} \\
  & \textcolor{nooverlap}{No Overlap}  & 76.73 & 42.67          & 62.66 / 51.43 & 7.45  / 0.59           \\

\midrule
\multirow{4}{*}{\makecell[l]{English--\\German}}
  & \textcolor{fulloverlap}{Full}      & 77.49 & \textbf{69.44} & 62.09 / 50.42 & \textbf{45.06 / 31.18} \\
  & \textcolor{highsim}{High-sim.}     & 78.40 & \textbf{69.98} & 62.24 / 51.43 & \textbf{45.32 / 32.52} \\
  & \textcolor{lowsim}{Low-sim.}       & 78.26 & \textbf{69.30} & 62.34 / 50.92 & \textbf{41.79 / 27.39} \\
  & \textcolor{nooverlap}{No Overlap}  & 78.08 & 35.01          & 61.96 / 49.83 & 5.09  / 0.25           \\

\midrule
\multirow{4}{*}{\makecell[l]{English--\\Turkish}}
  & \textcolor{fulloverlap}{Full}      & 77.54 & \textbf{49.46} & 61.03 / 49.75 & \textbf{22.16 / 11.85} \\
  & \textcolor{highsim}{High-sim.}     & \textbf{78.56} & \textbf{56.11} & 61.75 / 50.50 & \textbf{21.20 / 12.69} \\
  & \textcolor{lowsim}{Low-sim.}       & 78.40 & \textbf{52.32} & 62.02 / 51.01 & \textbf{20.41 / 11.60} \\
  & \textcolor{nooverlap}{No Overlap}  & 77.41 & 37.82          & 62.71 / 51.18 & 5.71  / 1.34           \\

\midrule
\multirow{4}{*}{\makecell[l]{English--\\Chinese}}
  & \textcolor{fulloverlap}{Full}      & \textbf{78.48} & \textbf{63.29} & 62.07 / 50.42 & \textbf{26.10 / 16.39} \\
  & \textcolor{highsim}{High-sim.}     & 77.15 & \textbf{60.42} & 62.75 / 50.50 & \textbf{23.56 / 16.30} \\
  & \textcolor{lowsim}{Low-sim.}       & 77.13 & \textbf{55.87} & 62.77 / 51.09 & \textbf{14.24 / 3.70}  \\
  & \textcolor{nooverlap}{No Overlap}  & 77.03 & 36.35          & 62.93 / 51.68 & 2.70  / 0.42           \\

\midrule
\multirow{4}{*}{\makecell[l]{English--\\Arabic}}
  & \textcolor{fulloverlap}{Full}      & 77.41 & \textbf{61.32} & 62.52 / 50.25 & \textbf{29.58 / 17.65}          \\
  & \textcolor{highsim}{High-sim.}     & 77.70 & \textbf{61.14} & 63.31 / 51.51 & \textbf{28.96 / 16.64}      \\
  & \textcolor{lowsim}{Low-sim.}       & 77.60 & \textbf{49.40} & 62.58 / 50.50 & \textbf{9.46 / 2.27}          \\
  & \textcolor{nooverlap}{No Overlap}  & 77.72 & 32.93          & 61.09 / 50.34 & 6.14 / 0.92              \\

\midrule
\multirow{4}{*}{\makecell[l]{English--\\Swahili}}
  & \textcolor{fulloverlap}{Full}      & 75.11 & \textbf{48.24} & \textemdash          & \textemdash                    \\
  & \textcolor{highsim}{High-sim.}     & 74.55 & \textbf{49.26} & \textemdash          & \textemdash                    \\
  & \textcolor{lowsim}{Low-sim.}       & 75.23 & \textbf{43.49} & \textemdash          & \textemdash                    \\
  & \textcolor{nooverlap}{No Overlap}  & 75.69 & 33.75          & \textemdash          & \textemdash                    \\

\bottomrule
\end{tabular}%
}
\caption{Downstream performance across language pairs and vocabulary overlap settings. For XNLI, we report accuracy; for XQuAD, we report F1 and exact match (EM). Settings significantly different from \emph{No Overlap} are in bold (see~\Cref{tab:mcnemar-all} for all $p$-values).
}
\label{tab:downstream}
\end{table}

\paragraph{Results.}
Results for both tasks are shown in \Cref{tab:downstream}. To compare XNLI accuracies and XQuAD exact match (EM) scores across models, we conducted pairwise McNemar tests (see \Cref{tab:mcnemar-all}). While $L_1$ (English) evaluation results are reported for completeness, we center the discussion here on $L_2$ transfer, which is the main focus of this work.

On $L_2$ transfer, the \textit{No Overlap} models performed substantially worse than all other overlap settings across every language pair for both downstream tasks (all $p < .05$). This confirms that some degree of shared vocabulary is always beneficial for cross-lingual transfer.
Comparisons between overlap types show more subtle patterns. \textit{Full Overlap} and \textit{High-similarity Overlap} achieved the strongest transfer performance overall: \textit{Full} was best in six of eleven $L_2$ evaluations (both XNLI accuracy and XQuAD F1/EM), while \textit{High-similarity} was best in four evaluations. However, differences between these two settings were not significant in seven of the eleven cases (all $p > .05$). By contrast, both \textit{Full} and \textit{High-similarity Overlap} consistently outperformed \textit{Low-similarity Overlap}: \textit{Full} was stronger in ten of eleven evaluations (seven significant; all $p < .05$), and \textit{High-similarity} was stronger in nine of eleven (seven significant; all $p < .05$). This advantage is notable given that high-similarity tokens make up only 10–20\% of training and evaluation corpora, whereas low-similarity tokens account for as much as 80\% (see~\Cref{app:overlap-metrics}). 
\looseness=-1

These results show that while any overlap helps, sharing semantically similar tokens is far more impactful. The language pair also matters: for languages closely related to English, such as Spanish and German, \textit{High-} and \textit{Low-similarity Overlap} performed comparably, whereas for more distant languages, \textit{High-similarity Overlap} gave a clearer advantage. In Chinese and Arabic, the use of a different script from English reduces the value of cross-lingual transfer in the \emph{Low-similarity Overlap} setting. Here, semantically aligned tokens have an outsized impact, as they are often English words introduced through code-switching. We provide the full list of overlapping tokens with their similarity scores in our repository.
\looseness=-1

\section{Conclusion}
In this paper, we present a detailed study of vocabulary overlap in multilingual language models. Our experimental design isolates the effect of overlap by controlling for token frequency and subword segmentation quality. We also uniquely disentangle how semantically similar or dissimilar vocabulary overlap affect multilingual representations and task transfer. While prior work has raised concerns that highly polysemantic tokens from vocabulary sharing may hinder performance,
we find that overlap (1) promotes alignment of the embedding spaces between languages in bilingual models and (2) enables cross-lingual transfer on downstream tasks. Overlapping tokens with the same meaning across languages contribute most, though any overlap proves beneficial. We therefore argue that, rather than reducing overlap, tokenizer development should focus on other determinants of quality, such as per-language compression rates.

\section{Acknowledgments}

The authors would like to thank Róbert Csordás, Tomasz Limisiewicz, Isabel Papadimitriou, and Ekaterina Shutova for helpful comments at different stages of the project. We would also like to thank the members of the Stanford NLP Group, the Jurafsky Lab Group, and the anonymous reviewers for useful discussions. Julie Kallini is supported by a National Science Foundation Graduate Research Fellowship under grant number DGE-2146755.

\section{Limitations}

Our study analyzes six language pairs spanning diverse language families. Each language pair requires pre-training four models, which is computationally expensive. While our selection of languages provides meaningful breadth, with additional compute resources, future work could extend the analysis to additional language pairs, particularly more low-resource languages. We also focus on English-centric pairs, reflecting common multilingual pre-training scenarios where English is the dominant language. Exploring overlap effects in non-English pairings would complement our findings.
In addition, we use a single, widely adopted tokenizer (XLM-R) to control for tokenizer quality across conditions. Although this choice allows for a clean comparison of overlap settings, future work could examine how overlap interacts with tokenizers of varying quality or design choices to further contextualize our results. Finally, following \citet{dufter-schutze-2020-identifying}, future work could explore whether extended training or different parameter budgets further affect cross-lingual generalization under the different overlap settings.

\bibliography{anthology,custom}

\newpage

\appendix
\onecolumn

\section{Tokenizer Compression Rates}
\label{app:tokenizer-compression}

\begin{figure}[ht]
  \centering
  \begin{subfigure}{0.9\columnwidth}
    \centering
    \includegraphics[width=\linewidth]{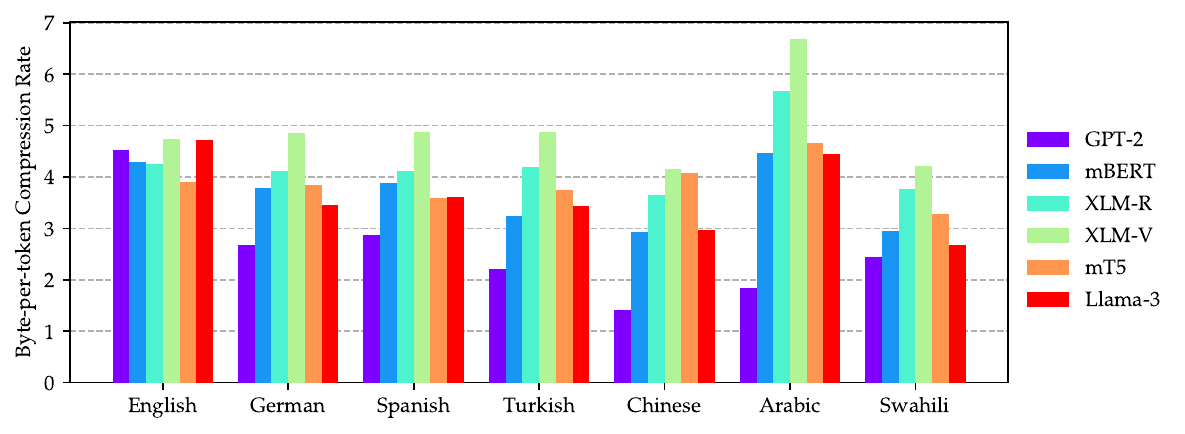}
    \caption{Byte-per-token compression rates.}
    \label{fig:byte-compression}
  \end{subfigure}

  \begin{subfigure}{0.9\columnwidth}
    \centering
    \includegraphics[width=\linewidth]{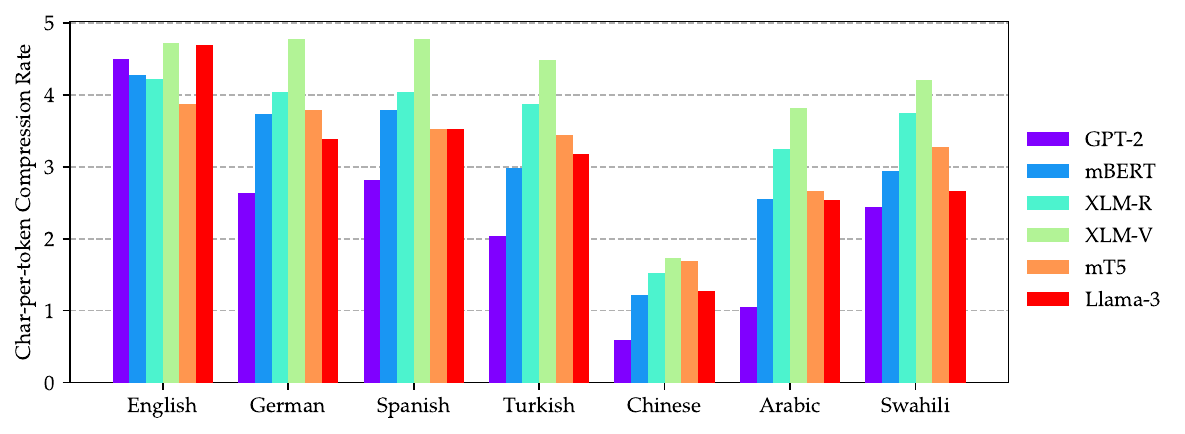}
    \caption{Character-per-token compression rates.}
    \label{fig:char-compression}
  \end{subfigure}

  \caption{Byte-per-token and character-per-token compression rates for English, German, Spanish, Turkish, Chinese, Arabic, and Swahili, for six different tokenizers.}
  \label{fig:tokenizer-compression}
\end{figure}

We consider six multilingual tokenizers as candidates for our base tokenizer $\mathcal{T}$: GPT-2 \cite{radford2019languagemodels}, mBERT \cite{devlin-etal-2019-bert}, XLM-R \cite{conneau-etal-2020-unsupervised}, XLM-V \cite{liang-etal-2023-xlm}, mT5 \cite{xue-etal-2021-mt5}, and Llama 3 \cite{grattafiori2024llama}. We compute byte-per-token and character-per-token compression rates for the seven languages involved in our study, using samples from the multilingual C4 corpus \cite{raffel2020exploring}. As shown in \Cref{fig:tokenizer-compression}, XLM-V achieves the best compression but has an extremely large vocabulary (1M tokens), making it impractical for our setup. XLM-R's compression is competitive with XLM-V's at a much smaller vocabulary size (250k tokens), making it a suitable choice for our controlled experiments.

\section{Layer Selection}
\label{app:layer-selection}

\begin{figure}[ht]
  \centering
   \includegraphics[width=0.6\linewidth]{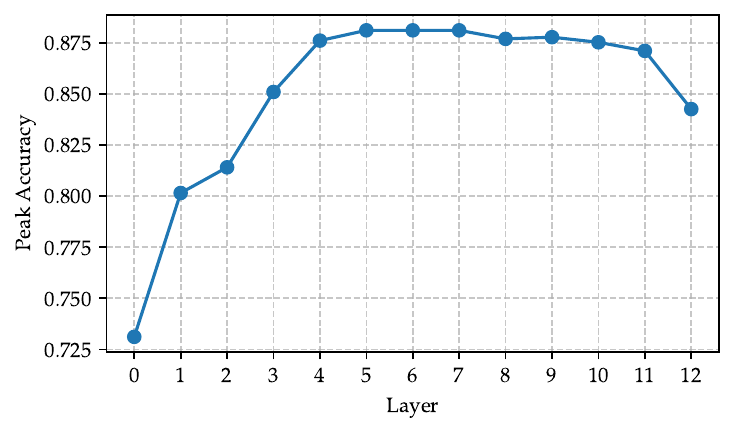}
  \caption{Results of our layer sweep on XLM-R using English--Dutch data from \citealt{lefever-etal-2020-identifying}.}
  \label{fig:layer-sweep}
\end{figure}

We select the Transformer layer that best distinguishes between semantically similar and dissimilar tokens in a controlled setup.
We use manually annotated data from the English--Dutch cognate detection dataset of \citet{lefever-etal-2020-identifying}, as well as English--Dutch parallel texts from CCMatrix.
From the cognate detection dataset, we extract a list of both cognates and non-cognates, which we tokenize using XLM-R's tokenizer.
We remove any words that are tokenized into more than one token, as well as non-overlapping tokens and tokens that appear fewer than 100 times in the parallel texts.
One author then manually verified that no cognates remained in the non-cognate subset.
For each remaining token, we sample 100 occurrences per language from the English--Dutch parallel texts.
We then pass these tokens through XLM-R's layers $l\in\{1,\dots,12\}$ and average the layer-$l$ embeddings to obtain a static embedding per token for each language.
We compute the cosine similarity between the static embeddings at each layer and rank the tokens by similarity.
To quantify the capacity of each layer to distinguish between cognates and non-cognates, we sweep through every possible threshold $n$ in the ranked list.
Specifically, we label the top-$n$ tokens as \textit{predicted cognates} and the remaining tokens as \textit{predicted non-cognates}, measuring the classification accuracy against our gold labels.
The highest classification accuracy over all $n$ is taken as the oracle score for layer $l$.
As shown in \Cref{fig:layer-sweep}, the highest classification accuracy was obtained using layer 5, and we therefore use layer 5 to rank tokens in $O$ by their semantic similarity across all language pairs.

\section{Overlap Metrics for All Datasets}
\label{app:overlap-metrics}

Below we report corpus statistics \emph{after} applying the four overlap manipulations described in~\Cref{sec:approach}.
For each language pair $L_1$ and $L_2$, we start with corpora $C_1$ and $C_2$ taken from CCMatrix and tokenize them with the XLM‑R SentencePiece tokenizer $\mathcal{T}$.
This yields individual language vocabulary sets $V_1=\{\text{unique tokens in }C_1\}$ and $V_2=\{\text{unique tokens in }C_2\}$. An overlap setting remaps token indices, producing new language vocabularies $V_1'$ and $V_2'$, with $|V_1'| = |V_1|$ and $|V_2'| = |V_2|$. Their intersection, $O' = V_1' \cap V_2'$ contains the tokens shared under that setting. Thus, the total effective vocabulary size of $\mathcal{T}'$ is $N_\mathrm{eff}' = |V_1| + |V_2| - |O'|$.
With these definitions in place, we now define two overlap metrics:

\begin{enumerate}
  \item \textbf{Type overlap (IoU).}  
        The Jaccard similarity of the setting‑specific vocabularies
        $V_1'$ and $V_2'$ is  
        $$
          J(V_1',V_2') \;=\;
          \frac{|V_1' \cap V_2'|}
               {|V_1' \cup V_2'|}
          \;=\;
          \frac{|O'|}
               {|V_1'|+|V_2'|-|O'|}
          \;=\;
          \frac{|O'|}{N_{\text{eff}}'}.
        $$

  \item \textbf{Frequency‑weighted overlap.}  
        To quantify how often the shared tokens are \emph{used} in each corpus,
        we compute, for $i\in\{1,2\}$,
        $$
          F_i \;=\;
          \frac{\sum_{t\in O'} \mathrm{count}_i(t)}
               {\sum_{t\in V_i'} \mathrm{count}_i(t)},
        $$
        where $\mathrm{count}_i(t)$ is the frequency of token $t$ in
        corpus $C_i'$, where $C_i'$ is the corpus $C_i$ after applying the token remapping under the given setting.  
        Thus $F_i$ is the proportion of running tokens in $C_i'$ that belong
        to the shared vocabulary $O'$.
\end{enumerate}
\Cref{tab:overlap-metrics} presents these statistics in the tokenized CCMatrix pre-training data for every language pair under each overlap condition. We also report frequency-weighted overlap metrics with respect to the XNLI and XQuAD training and test datasets in~\Cref{tab:downstream-freqs}. Remarkably, although the \textit{High‑} and \textit{Low‑similarity Overlap} settings contain the same number of overlapping token types, the latter has substantially higher frequency‑weighted overlaps in the pre-training corpora as well as the downstream task datasets.

\begin{table*}[ht]
\centering
\small
\label{tab:vocab-overlap}
\begin{tabular}{llccccccc}
\toprule
\textbf{Language Pair} & \textbf{Setting} & $|V_1|$ & $|V_2|$ & $|O'|$ & $N_{\mathrm{eff}}'$ & IoU (\%) & $F_1$ (\%) & $F_2$ (\%) \\
\midrule
\multirow{4}{*}{English--Spanish}
 & \textcolor{fulloverlap}{Full Overlap}        & \multirow{4}{*}{78,469} & \multirow{4}{*}{78,381} & 73,455 & 83,395  & 88.08 & 99.88 & 98.98\\
 & \textcolor{highsim}{High-sim.\ Overlap}      &                          &                          & 22,103 & 134,747 & 16.40 & 21.47 & 19.24\\
 & \textcolor{lowsim}{Low-sim.\ Overlap}        &                          &                          & 22,101 & 134,749 & 16.40 & 77.32 & 66.80\\
 & \textcolor{nooverlap}{No Overlap}            &                          &                          & 0      & 156,850 &  0.00 &  0.00 &  0.00\\
\midrule
\multirow{4}{*}{English--German}
 & \textcolor{fulloverlap}{Full Overlap}        & \multirow{4}{*}{83,126} & \multirow{4}{*}{83,884} & 75,922 & 91,088  & 83.35 & 96.73 & 99.06\\
 & \textcolor{highsim}{High-sim.\ Overlap}      &                          &                          & 20,594 & 146,416 & 14.07 & 20.37 & 18.68\\
 & \textcolor{lowsim}{Low-sim.\ Overlap}        &                          &                          & 20,592 & 146,418 & 14.06 & 75.82 & 68.05\\
 & \textcolor{nooverlap}{No Overlap}            &                          &                          & 0      & 167,010 &  0.00 &  0.00 &  0.00\\
\midrule
\multirow{4}{*}{English--Turkish}
 & \textcolor{fulloverlap}{Full Overlap}        & \multirow{4}{*}{65,665} & \multirow{4}{*}{69,703} & 58,724 & 76,644  & 76.62 & 99.99 & 86.58\\
 & \textcolor{highsim}{High-sim.\ Overlap}      &                          &                          & 13,906 & 121,462 & 11.45 & 19.92 & 17.20\\
 & \textcolor{lowsim}{Low-sim.\ Overlap}        &                          &                          & 13,907 & 121,461 & 11.45 & 76.81 & 43.03\\
 & \textcolor{nooverlap}{No Overlap}            &                          &                          & 0      & 135,368 &  0.00 &  0.00 &  0.00\\
\midrule
\multirow{4}{*}{English--Chinese}
 & \textcolor{fulloverlap}{Full Overlap}        & \multirow{4}{*}{67,754} & \multirow{4}{*}{73,491} & 57,102 & 84,143  & 67.86 & 99.99 & 71.09\\
 & \textcolor{highsim}{High-sim.\ Overlap}      &                          &                          & 12,598 & 128,647 &  9.79 & 22.59 &  9.26\\
 & \textcolor{lowsim}{Low-sim.\ Overlap}        &                          &                          & 12,599 & 128,646 &  9.79 & 73.04 & 19.20\\
 & \textcolor{nooverlap}{No Overlap}            &                          &                          & 0      & 141,245 &  0.00 &  0.00 &  0.00\\
\midrule
\multirow{4}{*}{English--Arabic}
 & \textcolor{fulloverlap}{Full Overlap}        & \multirow{4}{*}{69,129} & \multirow{4}{*}{68,975} & 57,084 & 81,020  & 70.46 & 96.11 & 61.02\\
 & \textcolor{highsim}{High-sim.\ Overlap}      &                          &                          & 9,963  & 128,141 &  7.78 & 20.39 &  9.87\\
 & \textcolor{lowsim}{Low-sim.\ Overlap}        &                          &                          & 9,963  & 128,141 &  7.78 & 67.56 &  8.19\\
 & \textcolor{nooverlap}{No Overlap}            &                          &                          & 0      & 138,104 &  0.00 &  0.00 &  0.00\\
\midrule
\multirow{4}{*}{English--Swahili}
 & \textcolor{fulloverlap}{Full Overlap}        & \multirow{4}{*}{45,699} & \multirow{4}{*}{41,956} & 37,275 & 50,380  & 73.99 & 97.67 & 79.55\\
 & \textcolor{highsim}{High-sim.\ Overlap}      &                          &                          & 4,733  & 82,922  &  5.71 & 20.44 & 17.35\\
 & \textcolor{lowsim}{Low-sim.\ Overlap}        &                          &                          & 4,734  & 82,921  &  5.71 & 51.36 & 39.90\\
 & \textcolor{nooverlap}{No Overlap}            &                          &                          & 0      & 87,655  &  0.00 &  0.00 &  0.00\\
\bottomrule
\end{tabular}
\caption{Token statistics for the CCMatrix pre-training corpora: native vocabulary sizes ($|V_1|$, $|V_2|$), overlap size ($|O'|$), the resulting effective vocabulary size ($N_\mathrm{eff}'$), and percentage-based overlap metrics (IoU, $F_1$, $F_2$) reported for every language pair and overlap setting.}
\label{tab:overlap-metrics}
\end{table*}

\begin{table*}[ht]
\centering
\small
\begin{tabular}{llcccccc}
\toprule
\multirow{2}{*}{\textbf{Language Pair}} & \multirow{2}{*}{\textbf{Setting}}
  & \multicolumn{3}{c}{\textbf{XNLI}} 
  & \multicolumn{3}{c}{\textbf{XQuAD}} \\ 
\cmidrule(lr){3-5} \cmidrule(lr){6-8}
  & & Train ($L_1$) & Test ($L_1$) & Test ($L_2$)
    & Train ($L_1$) & Test ($L_1$) & Test ($L_2$) \\
\midrule
\multirow{4}{*}{English--Spanish}
  & \textcolor{fulloverlap}{Full Overlap}       & 100.00 & 100.00 & 99.78
                                                &  99.99 & 100.00 &  99.79 \\
  & \textcolor{highsim}{High-sim.\ Overlap}     &  23.89 &  19.25 & 17.69
                                                &  19.85 &  19.67 &  16.29 \\
  & \textcolor{lowsim}{Low-sim.\ Overlap}       &  75.16 &  79.81 & 69.71
                                                &  79.06 &  79.19 &  72.20 \\
  & \textcolor{nooverlap}{No Overlap}           &   0.00 &   0.00 &  0.00
                                                &   0.00 &   0.00 &   0.00 \\
\midrule
\multirow{4}{*}{English--German}
  & \textcolor{fulloverlap}{Full Overlap}       & 100.00 & 100.00 & 99.50
                                                &  99.98 & 100.00 &  99.46 \\
  & \textcolor{highsim}{High-sim.\ Overlap}     &  22.23 &  16.85 & 15.67
                                                &  17.16 &  17.02 &  14.91 \\
  & \textcolor{lowsim}{Low-sim.\ Overlap}       &  77.26 &  82.65 & 71.09
                                                &  82.13 &  82.32 &  73.23 \\
  & \textcolor{nooverlap}{No Overlap}           &   0.00 &   0.00 &  0.00
                                                &   0.00 &   0.00 &   0.00 \\
\midrule
\multirow{4}{*}{English--Turkish}
  & \textcolor{fulloverlap}{Full Overlap}       &  99.99 &  99.99 & 86.35
                                                &  99.95 &  99.95 &  87.06 \\
  & \textcolor{highsim}{High-sim.\ Overlap}     &  22.97 &  17.48 & 14.90
                                                &  16.77 &  16.77 &  13.61 \\
  & \textcolor{lowsim}{Low-sim.\ Overlap}       &  73.49 &  78.62 & 43.74
                                                &  78.48 &  78.36 &  46.09 \\
  & \textcolor{nooverlap}{No Overlap}           &   0.00 &   0.00 &  0.00
                                                &   0.00 &   0.00 &   0.00 \\
\midrule
\multirow{4}{*}{English--Chinese}
  & \textcolor{fulloverlap}{Full Overlap}       &  99.98 &  99.99 & 71.08
                                                &  99.93 &  99.94 &  74.23 \\
  & \textcolor{highsim}{High-sim.\ Overlap}     &  21.92 &  22.38 &  8.47
                                                &  17.92 &  17.71 &   4.52 \\
  & \textcolor{lowsim}{Low-sim.\ Overlap}       &  73.46 &  72.85 & 18.35
                                                &  76.38 &  76.57 &  18.46 \\
  & \textcolor{nooverlap}{No Overlap}           &   0.00 &   0.00 &  0.00
                                                &   0.00 &   0.00 &   0.00 \\
\midrule
\multirow{4}{*}{English--Arabic}
  & \textcolor{fulloverlap}{Full Overlap}       &  99.96 &  99.95 & 61.19
                                                &  99.93 &  99.94 &  61.60 \\
  & \textcolor{highsim}{High-sim.\ Overlap}     &  21.52 &  21.54 & 10.39
                                                &  18.03 &  18.32 &   6.73 \\
  & \textcolor{lowsim}{Low-sim.\ Overlap}       &  70.08 &  69.84 &  7.46
                                                &  73.43 &  73.33 &   7.64 \\
  & \textcolor{nooverlap}{No Overlap}           &   0.00 &   0.00 &  0.00
                                                &   0.00 &   0.00 &   0.00 \\
\midrule
\multirow{4}{*}{English--Swahili}
  & \textcolor{fulloverlap}{Full Overlap}       &  97.56 &  97.34 & 79.13
                                                & \textemdash & \textemdash & \textemdash \\
  & \textcolor{highsim}{High-sim.\ Overlap}     &  22.81 &  18.04 & 16.03
                                                & \textemdash & \textemdash & \textemdash \\
  & \textcolor{lowsim}{Low-sim.\ Overlap}       &  48.01 &  51.34 & 41.00
                                                & \textemdash & \textemdash & \textemdash \\
  & \textcolor{nooverlap}{No Overlap}           &   0.00 &   0.00 &  0.00
                                                & \textemdash & \textemdash & \textemdash \\
\bottomrule
\end{tabular}%
\caption{Frequency-weighted overlap in the XNLI and XQuAD datasets for each language pair and vocabulary overlap setting. Higher values indicate a larger proportion of running tokens that come from the shared set $O'$.}
\label{tab:downstream-freqs}
\end{table*}

\section{Pre-training Experiment Details}
\label{app:pre-training}

\subsection{Model Architectures}
All of our models are autoregressive Transformers with a similar architecture to GPT-2 \cite{radford2019languagemodels} with 12 layers, 12 attention heads, $d_\mathrm{model} = 768$, and $d_\mathrm{ff} = 3072$. The only change we make to the standard GPT-2 architecture is the addition of rotary position embeddings (RoPE, \citealp{SU2024127063}), since this is the positional encoding method most often used in modern LLMs. The total non-embedding parameter count for all models is 85M, equivalent to the original GPT-2.

 To isolate the effect of vocabulary overlap, we tokenize the data once and vary only which tokens are shared, which necessarily results in different vocabulary sizes across settings. Thus, the total model parameters varies based on the setting and language pair.
To minimize unnecessary parameters, we prune the vocabulary to only retain tokens that appear in the CCMatrix corpus.
\Cref{tab:param_counts} reports the resulting vocabulary sizes and total parameter counts for every setting and language pair. For the English–Spanish and English–German pairs, the retained vocabularies are marginally larger than the effective sizes $N_\mathrm{eff}$ reported in~\Cref{tab:overlap-metrics}. This discrepancy occurs because the full CCMatrix corpora---on which the pruning was based---contain more tokens than the 6.6 billion tokens ultimately used for pre‑training; consequently, a small subset of the embedding matrix remained unused during training.

Here, we note that no single setting can claim an a priori advantage based solely on vocabulary size.
Larger vocabularies benefit from more model parameters but have higher upper bounds on perplexity and receive fewer gradient updates per embedding.

\subsection{Optimization}

We train with an effective batch size of 64 sequences, each 1024 tokens long, for a per‑step token count $2^{16}= 65,536$ tokens. The device batch size is 8 sequences. Each model is trained for a total of 100,000 gradient steps using the AdamW optimizer. The learning rate linearly warms up to $2.5\mathrm{e}{-4}$ during the first 5,000 steps, then follows a cosine decay.

Because the batch size and number of steps are identical across settings, each model processes 6.6 billion tokens in total. The required number of passes through CCMatrix therefore depends on the parallel corpus size: one epoch for English--Spanish and English--German, 2.1 epochs for English--Chinese, 3.6 epochs for English--Turkish; 2.4 epochs for English--Arabic; and 28.7 epochs for English--Swahili.

Each pre‑training job is executed on two NVIDIA RTX A6000 GPUs (48 GB), consuming approximately 96 GPU‑hours per model ($\approx$48 wall‑clock hours). Training the full suite of 24 models therefore required 48 GPUs and about 2304 GPU-hours in total.

\begin{table}[ht]
\centering
\small
\begin{tabular}{llcc}
\toprule
\textbf{Language Pair} & \textbf{Setting} & \textbf{Vocabulary Size} & \textbf{Total Parameters} \\
\midrule
\multirow{4}{*}{English--Spanish}
  & \textcolor{fulloverlap}{Full Overlap}        & 107,894 & 167.9M \\
  & \textcolor{highsim}{High-similarity Overlap} & 174,271 & 218.9M \\
  & \textcolor{lowsim}{Low-similarity Overlap}   & 174,271 & 218.9M \\
  & \textcolor{nooverlap}{No Overlap}            & 196,374 & 235.9M \\
\midrule
\multirow{4}{*}{English--German}
  & \textcolor{fulloverlap}{Full Overlap}        & 101,813 & 163.2M \\
  & \textcolor{highsim}{High-similarity Overlap} & 163,178 & 210.4M \\
  & \textcolor{lowsim}{Low-similarity Overlap}   & 163,178 & 210.4M \\
  & \textcolor{nooverlap}{No Overlap}            & 183,772 & 226.2M \\
\midrule
\multirow{4}{*}{English--Turkish}
  & \textcolor{fulloverlap}{Full Overlap}        & 76,645  & 143.9M \\
  & \textcolor{highsim}{High-similarity Overlap} & 121,463 & 178.3M \\
  & \textcolor{lowsim}{Low-similarity Overlap}   & 121,463 & 178.3M \\
  & \textcolor{nooverlap}{No Overlap}            & 135,370 & 189.0M \\
\midrule
\multirow{4}{*}{English--Chinese}
  & \textcolor{fulloverlap}{Full Overlap}        & 84,144  & 149.7M \\
  & \textcolor{highsim}{High-similarity Overlap} & 128,648 & 183.9M \\
  & \textcolor{lowsim}{Low-similarity Overlap}   & 128,648 & 183.9M \\
  & \textcolor{nooverlap}{No Overlap}            & 141,247 & 193.5M \\
\midrule
\multirow{4}{*}{English--Arabic}
  & \textcolor{fulloverlap}{Full Overlap}        & 81,020  & 147.3M \\
  & \textcolor{highsim}{High-similarity Overlap} & 128,142 & 183.5M \\
  & \textcolor{lowsim}{Low-similarity Overlap}   & 128,142 & 183.5M \\
  & \textcolor{nooverlap}{No Overlap}            & 138,106 & 191.1M \\
\midrule
\multirow{4}{*}{English--Swahili}
  & \textcolor{fulloverlap}{Full Overlap}        & 50,381  & 123.7M \\
  & \textcolor{highsim}{High-similarity Overlap} & 82,923  & 148.7M \\
  & \textcolor{lowsim}{Low-similarity Overlap}   & 82,923  & 148.7M \\
  & \textcolor{nooverlap}{No Overlap}            & 87,657  & 152.4M \\
\bottomrule
\end{tabular}
\caption{Vocabulary sizes and parameter counts for each overlap setting. Parameter counts are shown in millions (M).}
\label{tab:param_counts}
\end{table}

\section{Fine-tuning Experiment Details}
\label{app:fine-tuning}

For MultiNLI fine-tuning, we train each model for 5 epochs using a per-device batch size of 64 sequences and a maximum sequence length of 1024 tokens. Optimization is performed with AdamW using a cosine learning rate schedule without warmup. We conduct a hyperparameter sweep over three batch sizes (128, 256, 512) and three learning rates ($1\mathrm{e}{-5}$, $5\mathrm{e}{-5}$, $1\mathrm{e}{-4}$), saving checkpoints every 500 gradient steps. Because the number of epochs is fixed, the total number of steps varies with the batch size. This sweep is performed independently for each language pair and overlap setting, and we select the best model and checkpoint based on validation performance on MultiNLI. Each run is trained on a single NVIDIA RTX A6000 GPU (48GB) and takes approximately 1.5 GPU hours on average. Across 24 models and 9 hyperparameter configurations, the total compute cost is roughly 320 GPU hours.

For SQuAD fine-tuning, we train each model for 7 epochs with a per-device batch size of 16 sequences and a maximum sequence length of 1024 tokens. The optimizer settings and hyperparameter sweep configurations are the same used for MultiNLI, but we save checkpoints every 200 steps. Each run is also trained on a single NVIDIA RTX A6000 GPU (48GB) and takes about 2 GPU hours on average. Across 24 models and 9 hyperparameter configurations, the total compute amounts to roughly 430 GPU hours.

\section{Embedding Similarity Analysis Over Training}
\label{app:embedding-analysis-checkpoints}

\begin{figure}[ht]
  \centering
  \begin{subfigure}{\columnwidth}
    \centering
    \includegraphics[width=\linewidth]{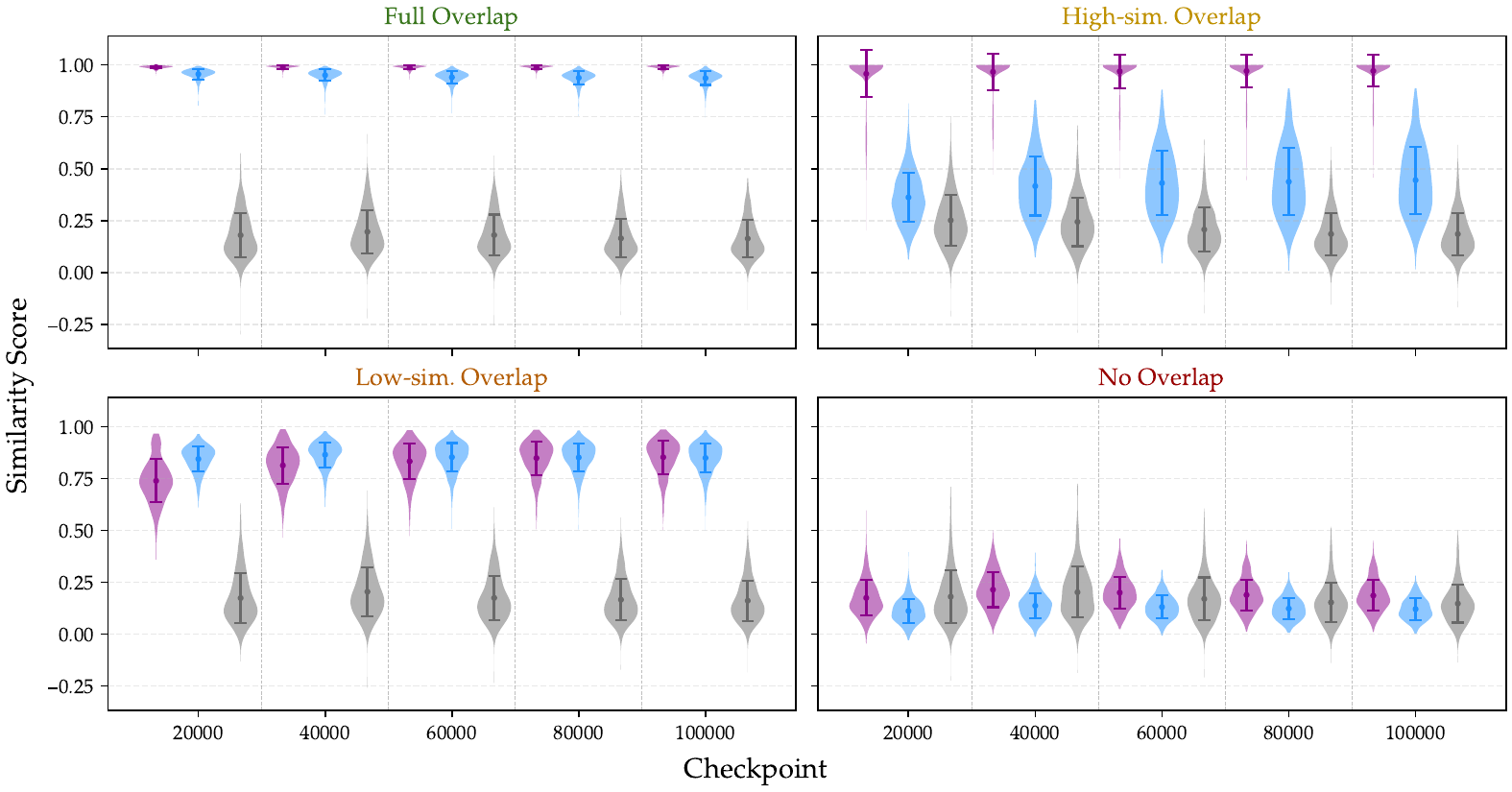}
    \caption{English--Spanish.}
    \label{fig:analysis-checkpoints-es}
  \end{subfigure}

  \begin{subfigure}{\columnwidth}
    \centering
    \includegraphics[width=\linewidth]{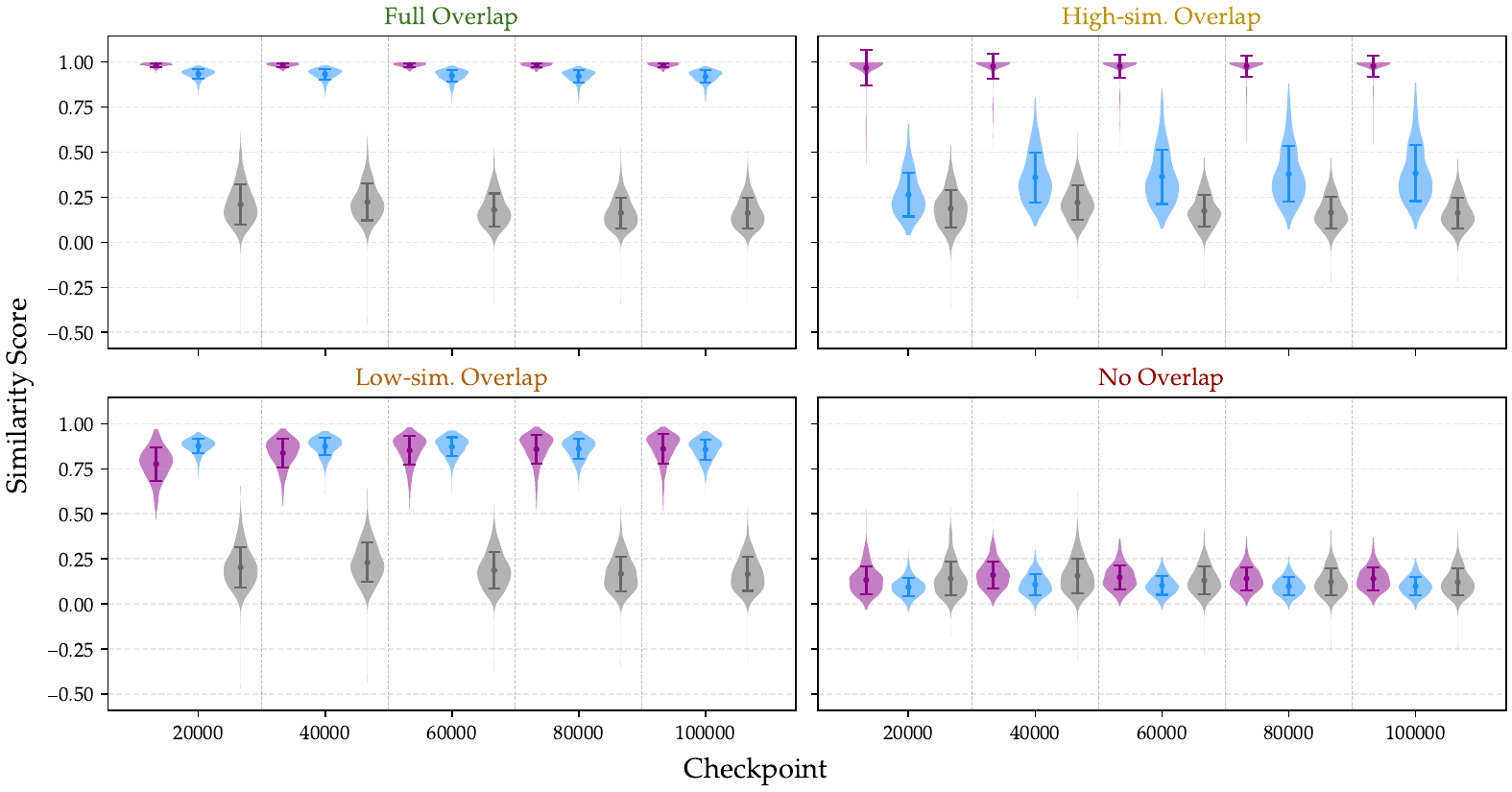}
    \caption{English--German.}
    \label{fig:analysis-checkpoints-de}
  \end{subfigure}

  \caption{Embedding similarity analysis for English--Spanish and English--German over pre-trained model checkpoints.}
  \label{fig:analysis-checkpoints-1}
\end{figure}

\begin{figure}[ht]
  \centering
  \begin{subfigure}{\columnwidth}
    \centering
    \includegraphics[width=\linewidth]{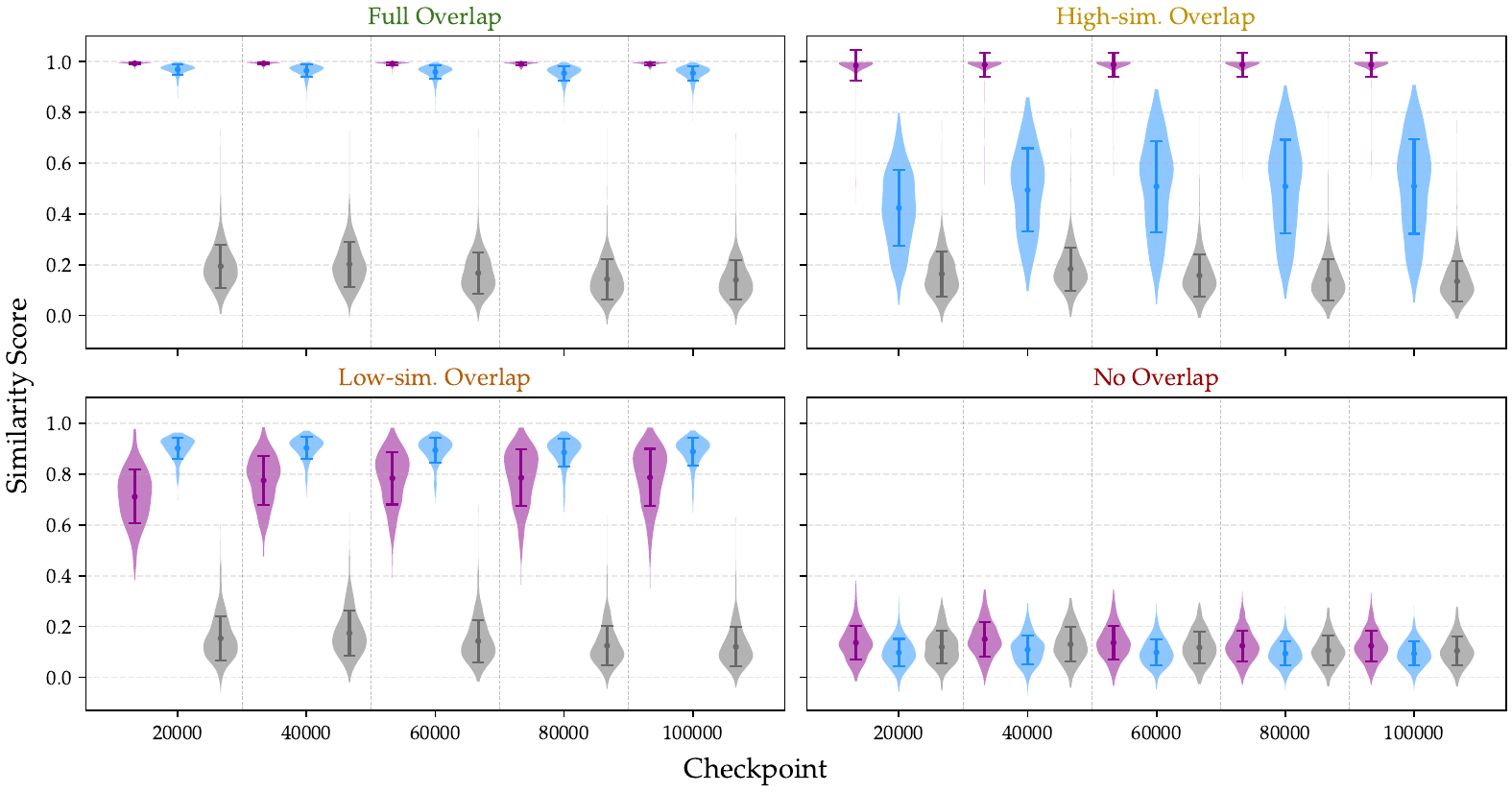}
    \caption{English--Turkish.}
    \label{fig:analysis-checkpoints-tr}
  \end{subfigure}

  \begin{subfigure}{\columnwidth}
    \centering
    \includegraphics[width=\linewidth]{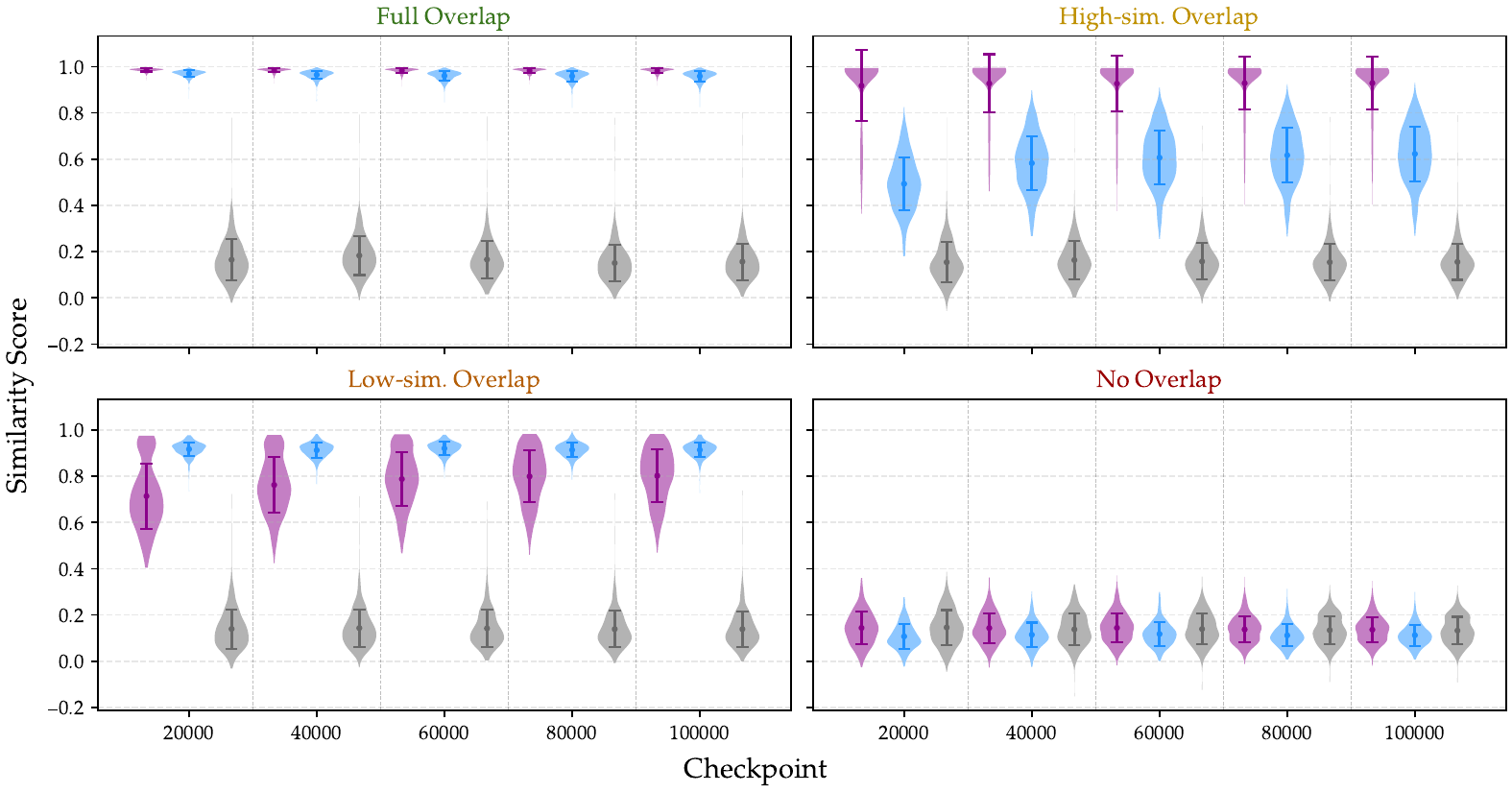}
    \caption{English--Chinese.}
    \label{fig:analysis-checkpoints-zh}
  \end{subfigure}

  \caption{Embedding similarity analysis for English--Turkish and English--Chinese over pre-trained model checkpoints.}
  \label{fig:analysis-checkpoints-2}
\end{figure}

\begin{figure}[ht]
  \centering
  \begin{subfigure}{\columnwidth}
    \centering
    \includegraphics[width=\linewidth]{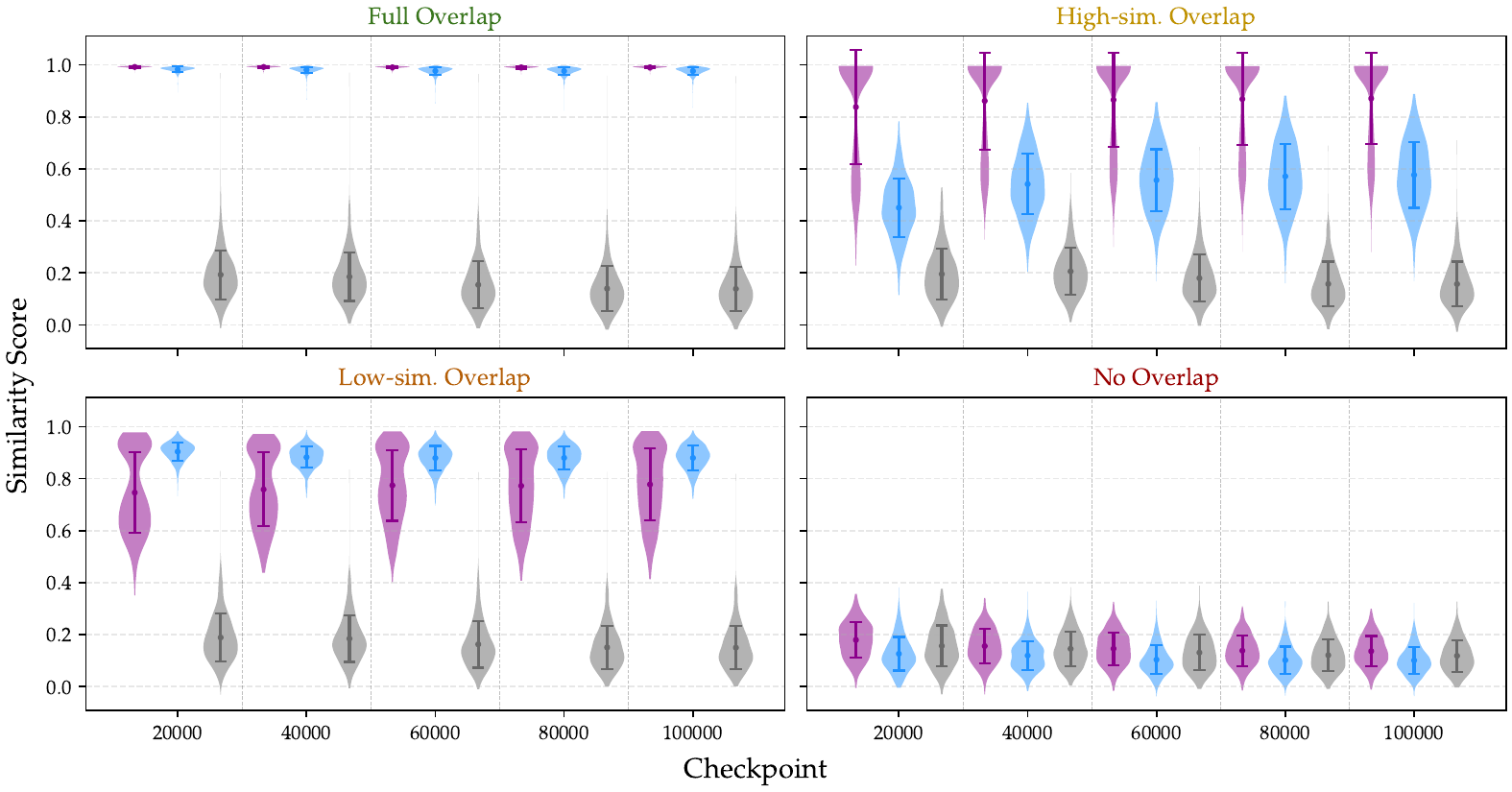}
    \caption{English--Arabic.}
    \label{fig:analysis-checkpoints-ar}
  \end{subfigure}

  \begin{subfigure}{\columnwidth}
    \centering
    \includegraphics[width=\linewidth]{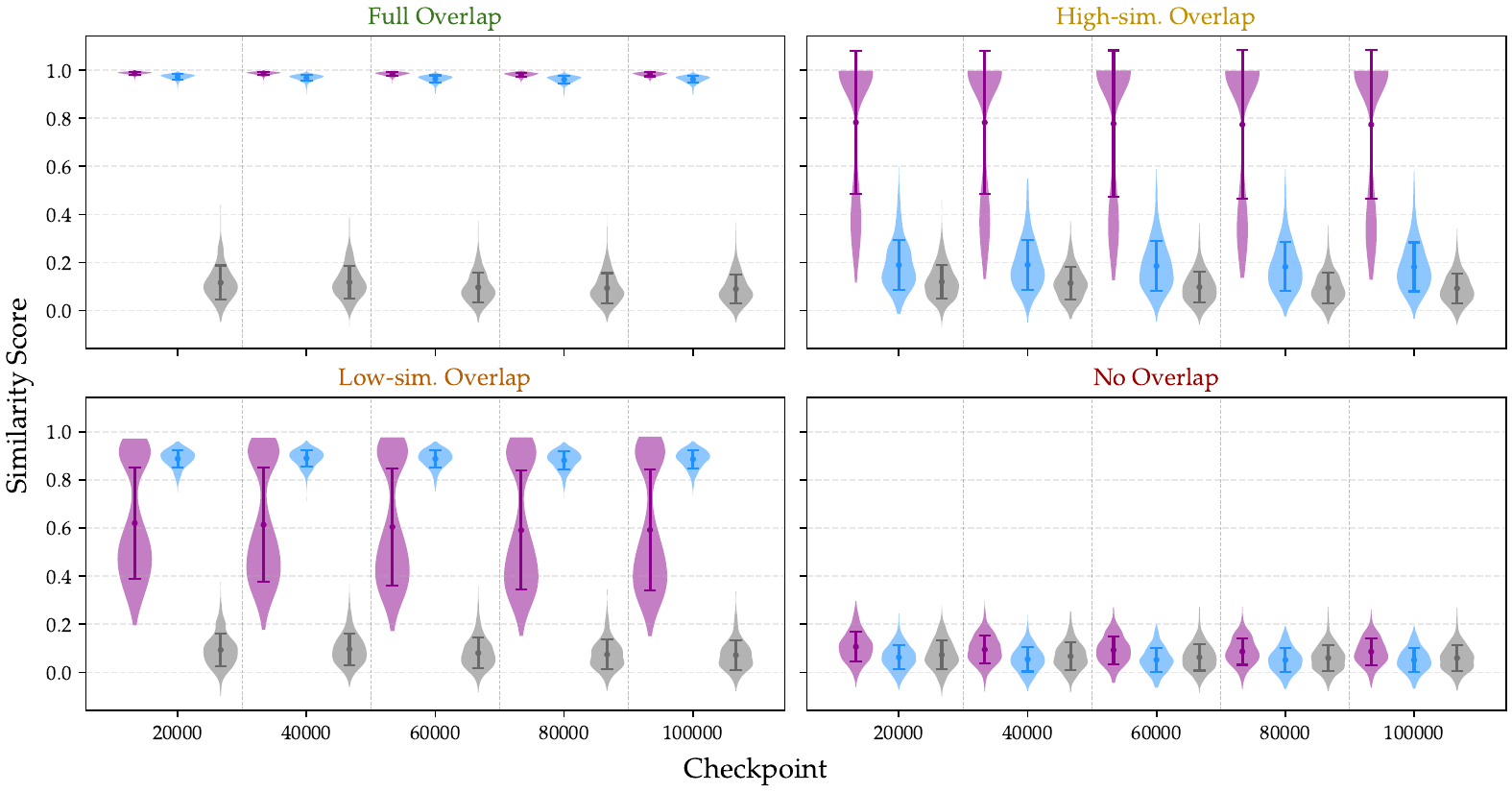}
    \caption{English--Swahili.}
    \label{fig:analysis-checkpoints-sw}
  \end{subfigure}

  \caption{Embedding similarity analysis for English--Arabic and English--Swahili over pre-trained model checkpoints.}
  \label{fig:analysis-checkpoints-3}
\end{figure}

In Figures~\ref{fig:analysis-checkpoints-1},~\ref{fig:analysis-checkpoints-2}, and~\ref{fig:analysis-checkpoints-3} we analyze embedding similarity at training checkpoints from 20k to 100k steps, in 20k increments. Across language pairs, we observe several trends. In the \emph{Full Overlap} setting, the scores for high-similarity tokens gradually separate from low-similarity ones over the course of training. \emph{High-similarity Overlap} shows a strong separation throughout training, with low-similarity tokens becoming more similar over time. In \emph{Low-similarity Overlap}, low-similarity tokens initially have higher similarity scores, but this reverses during training. \emph{No Overlap} shows little change in similarity scores over time.

\section{Significance Tests}

In this section, we present the Cohen's $d$ effect sizes for our embedding similarity analysis (\Cref{tab:cohens-d}), as well as the $p$-values for the pairwise McNemar tests between performance metrics on the XNLI and XQuAD downstream tasks (\Cref{tab:mcnemar-all}).

\begin{table}[h]
\centering
\small
\begin{tabular}{lcccc}
\toprule
\textbf{Language Pair} & \textcolor{fulloverlap}{\textbf{Full Overlap}} & \textcolor{highsim}{\textbf{High-Sim. Overlap}} & \textcolor{lowsim}{\textbf{Low-Sim. Overlap}} & \textcolor{nooverlap}{\textbf{No Overlap}} \\
\midrule
English--Spanish & 2.134 & 4.156 & 0.044 & 1.028 \\
English--German  & 2.458 & 5.053 & 0.049 & 0.721 \\
English--Turkish & 1.766 & 3.512 & -1.151 & 0.559 \\
English--Chinese & 1.358 & 2.642 & -1.350 & 0.467 \\
English--Arabic  & 1.264 & 1.918 & -0.992 & 0.646 \\
English--Swahili & 1.706 & 2.569 & -1.639 & 0.661 \\
\bottomrule
\end{tabular}
\caption{Cohen’s $d$ effect sizes from our embedding similarity analysis. These values compare the cosine similarities between the High-similarity and Low-similarity token sets for each language pair and vocabulary overlap condition.}
\label{tab:cohens-d}
\end{table}

\begin{table*}[ht]
  \centering

  \begin{subtable}{0.48\textwidth}
    \centering
    \small
    \resizebox{\textwidth}{!}{%
  \begin{tabular}{lrrr}
    \toprule
    \multicolumn{4}{c}{\bfseries English--Spanish ($L_1$)} \\
    Overlap Setting      & \textcolor{highsim}{High-sim. Overlap} & \textcolor{lowsim}{Low-sim. Overlap} & \textcolor{nooverlap}{No Overlap}  \\
    \midrule
\textcolor{fulloverlap}{Full Overlap}   & <.001 / .381  & <.001 / 1.000  & <.001 / .788 \\
\textcolor{highsim}{High-sim. Overlap}  & \textemdash    & .206 / .436   & .001 / .232 \\
\textcolor{lowsim}{Low-sim. Overlap}    & \textemdash    & \textemdash   & <.001 / .745 \\
    \midrule
    \multicolumn{4}{c}{\bfseries English--German ($L_1$)} \\
    Overlap Setting      & \textcolor{highsim}{High-sim. Overlap} & \textcolor{lowsim}{Low-sim. Overlap} & \textcolor{nooverlap}{No Overlap}  \\
    \midrule
\textcolor{fulloverlap}{Full Overlap}   & .076 / .454  & .130 / .734  & .253 / .675 \\
\textcolor{highsim}{High-sim. Overlap}  & \textemdash    & .812 / .708   & .551 / .211 \\
\textcolor{lowsim}{Low-sim. Overlap}    & \textemdash    & \textemdash   & .752 / .385 \\
    \midrule
    \multicolumn{4}{c}{\bfseries English--Turkish ($L_1$)} \\
    Overlap Setting      & \textcolor{highsim}{High-sim. Overlap} & \textcolor{lowsim}{Low-sim. Overlap} & \textcolor{nooverlap}{No Overlap}  \\
    \midrule
\textcolor{fulloverlap}{Full Overlap}   & .039 / .571  & .097 / .365  & .810 / .307 \\
\textcolor{highsim}{High-sim. Overlap}  & \textemdash    & .781 / .725   & .023 / .631 \\
\textcolor{lowsim}{Low-sim. Overlap}    & \textemdash    & \textemdash   & .062 / .945 \\
    \midrule
    \multicolumn{4}{c}{\bfseries English--Chinese ($L_1$)} \\
    Overlap Setting      & \textcolor{highsim}{High-sim. Overlap} & \textcolor{lowsim}{Low-sim. Overlap} & \textcolor{nooverlap}{No Overlap}  \\
    \midrule
\textcolor{fulloverlap}{Full Overlap}   & .011 / 1.000  & .012 / .640  & .005 / .340 \\
\textcolor{highsim}{High-sim. Overlap}  & \textemdash    & 1.000 / .688   & .844 / .393 \\
\textcolor{lowsim}{Low-sim. Overlap}    & \textemdash    & \textemdash   & .879 / .687 \\
    \midrule
    \multicolumn{4}{c}{\bfseries English--Arabic ($L_1$)} \\
    Overlap Setting      & \textcolor{highsim}{High-sim. Overlap} & \textcolor{lowsim}{Low-sim. Overlap} & \textcolor{nooverlap}{No Overlap}  \\
    \midrule
\textcolor{fulloverlap}{Full Overlap}   & .592 / .337  & .730 / .890  & .574 / 1.000 \\
\textcolor{highsim}{High-sim. Overlap}  & \textemdash    & .871 / .456   & 1.000 / .401 \\
\textcolor{lowsim}{Low-sim. Overlap}    & \textemdash    & \textemdash   & .850 / .947 \\
    \midrule
    \multicolumn{4}{c}{\bfseries English--Swahili ($L_1$)} \\
    Overlap Setting      & \textcolor{highsim}{High-sim. Overlap} & \textcolor{lowsim}{Low-sim. Overlap} & \textcolor{nooverlap}{No Overlap}  \\
    \midrule
\textcolor{fulloverlap}{Full Overlap}   & .313 / \textemdash  & .853 / \textemdash  & .291 / \textemdash \\
\textcolor{highsim}{High-sim. Overlap}  & \textemdash    & .235 / \textemdash   & .038 / \textemdash \\
\textcolor{lowsim}{Low-sim. Overlap}    & \textemdash    & \textemdash   & .413 / \textemdash \\
    \bottomrule
  \end{tabular}
  }
    \caption{$L_1$ (English) results.}
    \label{tab:l1-mcnemar}
  \end{subtable}
  \hfill
  \begin{subtable}{0.488\textwidth}
    \centering
    \small
    \resizebox{\textwidth}{!}{%
  \begin{tabular}{lrrr}
    \toprule
    \multicolumn{4}{c}{\bfseries English--Spanish ($L_2$)} \\
    Overlap Setting      & \textcolor{highsim}{High-sim. Overlap} & \textcolor{lowsim}{Low-sim. Overlap} & \textcolor{nooverlap}{No Overlap}  \\
    \midrule
\textcolor{fulloverlap}{Full Overlap}   & <.001 / <.001  & <.001 / .841  & <.001 / <.001 \\
\textcolor{highsim}{High-sim. Overlap}  & \textemdash    & .322 / .002   & <.001 / <.001 \\
\textcolor{lowsim}{Low-sim. Overlap}    & \textemdash    & \textemdash   & <.001 / <.001 \\
    \midrule
    \multicolumn{4}{c}{\bfseries English--German ($L_2$)} \\
    Overlap Setting      & \textcolor{highsim}{High-sim. Overlap} & \textcolor{lowsim}{Low-sim. Overlap} & \textcolor{nooverlap}{No Overlap}  \\
    \midrule
\textcolor{fulloverlap}{Full Overlap}   & .366 / .305    & .837 / .002  & <.001 / <.001 \\
\textcolor{highsim}{High-sim. Overlap}  & \textemdash    & .277 / <.001 & <.001 / <.001 \\
\textcolor{lowsim}{Low-sim. Overlap}    & \textemdash    & \textemdash  & <.001 / <.001 \\
    \midrule
    \multicolumn{4}{c}{\bfseries English--Turkish ($L_2$)} \\
    Overlap Setting      & \textcolor{highsim}{High-sim. Overlap} & \textcolor{lowsim}{Low-sim. Overlap} & \textcolor{nooverlap}{No Overlap}  \\
    \midrule
\textcolor{fulloverlap}{Full Overlap}   & <.001 / .423  & <.001 / .867  & <.001 / <.001 \\
\textcolor{highsim}{High-sim. Overlap}  & \textemdash  & <.001 / .319  & <.001 / <.001 \\
\textcolor{lowsim}{Low-sim. Overlap}    & \textemdash  & \textemdash  & <.001 / <.001 \\
    \midrule
    \multicolumn{4}{c}{\bfseries English--Chinese ($L_2$)} \\
    Overlap Setting      & \textcolor{highsim}{High-sim. Overlap} & \textcolor{lowsim}{Low-sim. Overlap} & \textcolor{nooverlap}{No Overlap}  \\
    \midrule
\textcolor{fulloverlap}{Full Overlap}    & <.001 / 1.000  & <.001 / <.001  & <.001 / <.001 \\
\textcolor{highsim}{High-sim. Overlap}   & \textemdash        & <.001 / <.001  & <.001 / <.001 \\
\textcolor{lowsim}{Low-sim. Overlap}     & \textemdash        & \textemdash  & <.001 / <.001 \\
    \midrule
    \multicolumn{4}{c}{\bfseries English--Arabic ($L_2$)} \\
    Overlap Setting      & \textcolor{highsim}{High-sim. Overlap} & \textcolor{lowsim}{Low-sim. Overlap} & \textcolor{nooverlap}{No Overlap}  \\
    \midrule
\textcolor{fulloverlap}{Full Overlap}    & .818 / .425  & <.001 / <.001  & <.001 / <.001 \\
\textcolor{highsim}{High-sim. Overlap}   & \textemdash        & <.001 / <.001  & <.001 / <.001 \\
\textcolor{lowsim}{Low-sim. Overlap}     & \textemdash        & \textemdash  & <.001 / .008 \\
    \midrule
    \multicolumn{4}{c}{\bfseries English--Swahili ($L_2$)} \\
    Overlap Setting      & \textcolor{highsim}{High-sim. Overlap} & \textcolor{lowsim}{Low-sim. Overlap} & \textcolor{nooverlap}{No Overlap}  \\
    \midrule
\textcolor{fulloverlap}{Full Overlap}    & .208 / \textemdash  & <.001 / \textemdash   & <.001 / \textemdash  \\
\textcolor{highsim}{High-sim. Overlap}   & \textemdash        & <.001 / \textemdash   & <.001 / \textemdash  \\
\textcolor{lowsim}{Low-sim. Overlap}     & \textemdash        & \textemdash  & <.001 / \textemdash  \\
    \bottomrule
  \end{tabular}
  }
    \caption{$L_2$ transfer results.}
    \label{tab:l2-mcnemar}
  \end{subtable}

  \caption{McNemar $p$‐values for XNLI / XQuAD across all overlap settings and language pairs. (a) presents results on $L_1$ (English); (b) presents $L_2$ transfer results. In each table entry, the first number is XNLI; the second is XQuAD.}
  \label{tab:mcnemar-all}
\end{table*}

\section{Licenses}

The CCMatrix corpus was released under the BSD license, and XLM-R was released under the MIT license. We will release our code and models under the MIT license. Our use of these artifacts is consistent with their intended use.

\section{Software Packages}

We use the following software libraries in our experiments: HuggingFace Transformers v4.47.0, Datasets v3.2.0, PyTorch v2.5.1, SentencePiece v0.2.0, and Statsmodels v0.14.4.

\end{document}